%% file: main.tex
\documentclass[conference]{IEEEtran}
\IEEEoverridecommandlockouts
\usepackage{cite}
\usepackage{amsmath,amssymb,amsfonts}
\usepackage{algorithmic}
\usepackage{graphicx}
\usepackage{textcomp}
\usepackage{xcolor}

\usepackage{microtype}
\usepackage{booktabs} 

\usepackage{hyperref}

\usepackage{multirow}
\usepackage{xspace}

\input{acrom}
\usepackage[capitalise]{cleveref}

\usepackage{fancyhdr}

\def\BibTeX{{\rm B\kern-.05em{\sc i\kern-.025em b}\kern-.08em
    T\kern-.1667em\lower.7ex\hbox{E}\kern-.125emX}}
\begin{document}
\setlength{\columnsep}{0.3in}



\title{\footnotesize
\framebox[1.01\width]{\parbox{\dimexpr\linewidth-2\fboxsep-2\fboxrule}{If you cite this paper, please use the ISIoT 2026 reference: Z. Huang, K. Schleiser, G. Myung, and E. Baccelli. Ariel-ML: Embedded Rust Leveraging Multicore for Neural Networks on Heterogeneous Microcontrollers. in Proceedings of the 8th International Workshop on Intelligent Systems for the Internet of Things (ISIoT 2026).}}
 \ \\ \ \\ \ \\
\Huge Ariel-ML: Embedded Rust Leveraging Multicore for Neural Networks on Heterogeneous Microcontrollers}

\author{
\IEEEauthorblockN{Zhaolan Huang\IEEEauthorrefmark{1}, Kaspar Schleiser\IEEEauthorrefmark{1}, Gyungmin Myung\IEEEauthorrefmark{2}, Emmanuel Baccelli\IEEEauthorrefmark{1}\IEEEauthorrefmark{3}}
\IEEEauthorblockA{\IEEEauthorrefmark{1} Freie Universit\"at Berlin, Germany \ \\
\IEEEauthorrefmark{2} KAIST, South Korea \ \\
\IEEEauthorrefmark{3} Inria, France
}
}

\maketitle

\begin{abstract}
\input{sections/00-abstract}
\end{abstract}


\input{sections/01-intro}
\input{sections/02-contribs}
\input{sections/03-background}

\input{sections/04-design}

\input{sections/05-implem}
\input{sections/06-benchmarks}
\input{sections/07-discussion}
\input{sections/08-related-work}
\input{sections/09-conclusion}

\bibliographystyle{IEEEtran}
\bibliography{bibliography}

\end{document}

%% file: acrom.tex
\usepackage{acronym}
\newacro{AI}{Artificial Intelligence}
\newacro{DAG}{directed acyclic graph}
\newacro{FL}{Federated Learning}
\newacro{ML}{Machine Learning}
\newacro{NN}{Neural Network}
\newacro{CNN}{Convolutional Neural Network}
\newacro{MPU}{memory protection unit}
\newacro{TMLaaS}{TinyML-as-a-Service}
\newacro{IoT}{Internet of Things}
\newacro{CBOR}{Concise Binary Object Representation}
\newacro{CoAP}{Constrained Application Protocol}
\newacro{IID} {independent and identically distributed} 
\newacro{MCU}{Microcontroller Unit}
\newacro{I$^2$C}{Inter-Integrated Circuit}
\newacro{IETF}{Internet Engineering Task Force}
\newacro{MLP}{Multi-Layer Perceptron} 
\newacro{ANN}{Artifical Neural Networks}
\newacro{TinyML}{Tiny Machine Learning}
\newacro{AIoT}{Artificial Intelligence of Things}
\newacro{ROC}{Receiver Operating Characteristic}
\newacro{AUC}{Area Under the Curve}
\newacro{WASN}{Wireless acoustic sensor network}
\newacro{ASN}{Acoustic sensor network}
\newacro{PTQ}{Post-Training Quantization}
\newacro{TPR}{True Positive Rate}
\newacro{FPR}{False Positive Rate}
\newacro{STFT}{Short-Time Fourier Transform}
\newacro{SNR}{Signal-to-noise ratio}
\newacro{DC}{Direct Current}
\newacro{OOM}{out-of-memory}
\newacro{NAS}{Neural Architecture Search}
\newacro{TVM}{Tensor Virtual Machine}
\newacro{DNN}{Deep Neural Network}
\newacro{MAC}{multiply–accumulate}
\newacro{RAM}{Random Access Memory}
\newacro{IR}{intermediate representation}
\newacro{SOTA}{state-of-the-art}
\newacro{DRAM}{Dynamic RAM}
\newacro{FPGA}{Field Programmable Gate Array}
\newacro{Conv1D}{1-D convolution}
\newacro{TCN}{Temporal Convolutional Network}
\newacro{RNN}{Recurrent Neural Network}
\newacro{SSM}{State-Space Model}
\newacro{GTA}{Global Temporal Aggregator}
\newacro{LSTM}{Long Short-Term Memory}
\newacro{GRU}{Gated Recurrent Unit}
\newacro{EEG}{electroencephalogram}
\newacro{SIMD}{Single Instruction Multiple Data}
\newacro{OTA}{Over-the-Air}

%% file: sections/00-abstract.tex
Low-power microcontroller (MCU) hardware is currently evolving from single-core architectures to predominantly multi-core architectures. In parallel, new embedded software building blocks are more and more written in Rust, while C/C++ dominance fades in this domain. On the other hand, small artificial neural networks (ANN) of various kinds are increasingly deployed in edge AI use cases, thus deployed and executed directly on low-power MCUs. In this context, both incremental improvements and novel innovative services will have to be continuously retrofitted using ANNs execution in software embedded on sensing/actuating systems already deployed in the field. However, there was so far no Rust embedded software platform automating parallelization for inference computation on multi-core MCUs executing arbitrary TinyML models. As new microcontroller hardware architectures are increasingly multicore, this gap must be bridged. This paper thus fills this gap by introducing Ariel-ML, a novel toolkit we designed combining a generic TinyML pipeline and an embedded Rust software platform which can take full advantage of multi-core capabilities of various 32-bit microcontroller families (Arm Cortex-M, RISC-V, ESP-32). We published the full open source code of its implementation, which we used to benchmark its capabilities using a zoo of various TinyML models. We show that Ariel-ML outperforms prior art in terms of inference latency as expected, and we show that, compared to pre-existing toolkits using embedded C/C++, Ariel-ML achieves comparable memory footprints. Ariel-ML thus provides a useful basis for TinyML practitioners and resource-constrained embedded Rust.

%% file: sections/01-intro.tex
\section{Introduction}

Running \ac{AI} directly on the smallest networked devices yields specific challenges that are tackled by a research domain sometimes coined \ac{AIoT}~\cite{ghosh2018artificial}, edge \ac{AI} or \ac{TinyML}.
In essence, \ac{AIoT} pushes the miniaturization of artificial neural networks so as to fit the resource constraints of microcontroller-based hardware. To measure the challenge, we can refer on the one hand to RFC7228~\cite{rfc7228}, which categorizes billions of \ac{IoT} devices running with total \ac{RAM} smaller than 100 KiB, Flash memory smaller than 500 KiB, and CPU clock speeds in MHz. On the other hand, even a single convolutional layer in a common (small) model such as  ResNet-34~\cite{koonce2021resnet-34, he2016deepresnet} requires more than 400 KiB in \ac{RAM} even after quantization. This example highlights the substantial gap that this domain must address.

While memory budgets do not fundamentally change on microcontroller unit (MCU) hardware due to energy consumption and cost constraints, newly shipped \ac{MCU}  hardware (ARM Cortex-M, RISC-V 32-bit, Expressif ESP32) is currently evolving from single-core architectures to predominantly multi-core architectures. In the context of \ac{AIoT}, efficiently exploiting multiple cores, on-demand at crunch time (e.g. for inference) becomes necessary to keep within real-time constraints, while employing optimizations decreasing the memory footprint.

In parallel, as our digital infrastructure becomes more critical in our way of life and, simultaneously, more under attack from a variety of actors (profit- or geopolitically-motivated), the need for novel, more secure systems software building blocks becomes a top priority. In the EU, the Cyber Resilience Act~\cite{eu-cra} is significantly stepping up security requirements for software building blocks, starting 2027.  In the USA, the White House was also calling for more secure foundations for systems software~\cite{white-house-report}. In this context, there is a huge push for alternative, developer-friendly programming languages offering a safer foundation than C/C++ which has been the dominant language for core systems software so far (e.g., for Linux or RIOT). One example of this push is the new code merged in Android: since 2025, the number of new lines in Rust has taken over the number of new lines in C/C++~\cite{android-rust}.

To accelerate this domain, the need emerges for a generic \ac{TinyML} pipeline combined with a general-purpose embedded \ac{IoT}  operating system, supporting out-of-the-box most microcontrollers, sensors and actuators, with a wide variety \ac{TinyML} models. The closest related work in that regard is probably RIOT-ML~\cite{huang2025riot-ml} and U-TOE~\cite{huang2023u}. However, these pipelines and their embedded software platform do not support parallel computation of inference on multi-core microcontrollers. Moreover, their embedded software platform is primarily implemented in C/C++. On the other hand, existing Rust machine learning frameworks such as Burn~\cite{burn-rs} or Candle~\cite{candle-rs} are neither tailored for low-power microcontroller targets, nor integrated in any Rust embedded \ac{IoT} software platform.  

To the best of our knowledge, there was so far no Rust embedded \ac{IoT}  software platform automating parallelization for inference computation on various multi-core \acp{MCU} executing arbitrary \ac{TinyML} models. This paper thus described work we have done aiming to fill this gap.

%% file: sections/02-contribs.tex
\subsection{Contributions}

\begin{itemize}
    \item We introduce Ariel-ML, a novel toolkit we designed for universal on-board evaluation of \ac{TinyML} models on low-power devices using a small embedded Rust runtime. In particular, Ariel-ML is the first such toolkit to natively support multi-core capabilities on microcontrollers. 
  
    \item We published the source code
    of Ariel-ML under an open source license. This implementation enables cross-compilation, automated multi-core optimization, flashing, and running (inference) with various neural networks (computational graph-based models) from mainstream \ac{ML} frameworks onto various low-power boards based on popular 32-bit microcontroller instruction set architectures (ARM Cortex-M, RISC-V).

    \item We provide benchmarks and a comparative experimental evaluation using Ariel-ML, reproducible both on an open-access \ac{IoT} testbed and on personal workstations, which provide insights on inference performance with different models on different low-power hardware and demonstrate how Ariel-ML can be reused by \ac{TinyML} experimental researchers and developers to fine-tune \ac{IoT}  configurations. 

\end{itemize}

%% file: sections/03-background.tex
\section{Background}

\subsection{Parallelism in Neural Network}

Neural networks have benefited from parallel computing for decades. Their core operations, convolution and matrix multiplication, are inherently parallelizable, making it straightforward to design conflict-free and efficient computation schedules. Meanwhile, the increasing computational demands of neural applications have driven hardware–software co-design, along with the evolution of high-performance computing and hardware accelerators.

Historically, the applications of neural networks were constrained by the scarcity of computational resources. Early studies focused on the use of parallel instructions (such as \ac{SIMD}) to improve single-core throughput \cite{vanhoucke2011improving} and explored various multi-core scheduling strategies to release the potential of many-core processors \cite{chu2006map}. The emergence of general-purpose GPU (GPGPU) computing revolutionized the field by developing fine-grained parallel operations over large tensors, enabling the large-scale neural network training and inference \cite{luo2005artificial}.

Nevertheless, achieving efficient and scalable parallelism remains a challenge, particularly in the context of \ac{IoT} systems, where multicore computing is still in its infancy. Recent work, such as TinyFormer \cite{jung2024tinyformer}, has demonstrated the potential to accelerate Transformer models on RISC-V–based multicore \ac{IoT} platforms. However, its \ac{MCU}-specific compilation flow and tightly coupled scheduling mechanism restrict its applicability to a wider range of microcontrollers, underscoring the need for more general and flexible parallelization frameworks.

\subsection{Multi-Core Scheduling on MCUs}
Scheduling assigns processes (\textit{threads}) to the available processors (\emph{cores}). Threads waiting to be scheduled are listed/sorted in a \emph{runqueue}.
The main challenges are to avoid idle cores, race conditions, or priority inversion, and favor low-overhead for runqueue maintenance.
The main examples of multicore scheduling on microcontrollers are the schedulers used in various real-time operating systems (\textit{RTOS}). These include, for instance, ThreadX~\cite{threadx}, FreeRTOS~\cite{freertos}, Zephyr~\cite{zephyr}, NuttX~\cite{nuttx}, or Ariel OS~\cite{ariel-os}.
Such schedulers typically use an aggregated runqueue (so-called \textit{global scheduling}), and global critical sections for synchronization within the scheduler.
To determine which core a thread should be allocated, ThreadX uses a reallocation routine that maps and balances the \textit{n} highest priority threads to the \textit{N} cores. 
In contrast, Ariel OS, FreeRTOS, Zephyr or NuttX use an approach whereby a core's next thread allocation is selected directly from the runqueue by the scheduler interrupt handler upon invocation. To the best of our knowledge, Ariel OS is so far the only embedded Rust RTOS that supports multicore at this level.

\subsection{IREE for Parallelism and Cross-Compilation}
Modern \ac{ML} compilers make parallelism patterns portable across heterogeneous hardware platforms. IREE (Intermediate Representation Execution Environment) \cite{The_IREE_Authors_IREE_2019} is an extensible MLIR-based compiler and runtime that lowers models from common frontends to modular backends for CPUs, GPUs (via LLVM, Vulkan, CUDA/Metal), and custom accelerators. Its multi-level IR and HAL abstractions allow parallel constructs such as tiled kernels, vectorized loops, and asynchronous dispatches to be expressed once and mapped to different devices by changing the backend, rather than redesigning the parallelization scheme for each target. This cross-compilation model, in which a single model artifact is reused across architectures and operating systems, aligns with Rust’s ethos of “fearless” cross-compilation. Previous work, such as TinyIREE~\cite{Liu2022TinyIREE}, shows that this approach extends to MCU-class systems, which we leverage in Ariel-ML by compiling models once with IREE and executing them on our Rust-based Ariel OS.

\subsection{Performance Metrics}
The bottleneck in all our use cases is obviously the constraint on resources available on the microcontroller(s) executing inference, using a given artificial neural network. Thus, for each model in our zoo, we focus only on what the pipeline deploys on the microcontroller. In detail, for a selection of typical models, we measure performance based on: 
\begin{itemize}
    \item Flash memory footprint;
    \item peak RAM memory use on the device;
    \item inference execution time (latency) on the device.
\end{itemize}.

%% file: sections/04-design.tex
\section{Ariel-ML Design}

\begin{figure}[h]
    \centering
    \includegraphics[width=\linewidth]{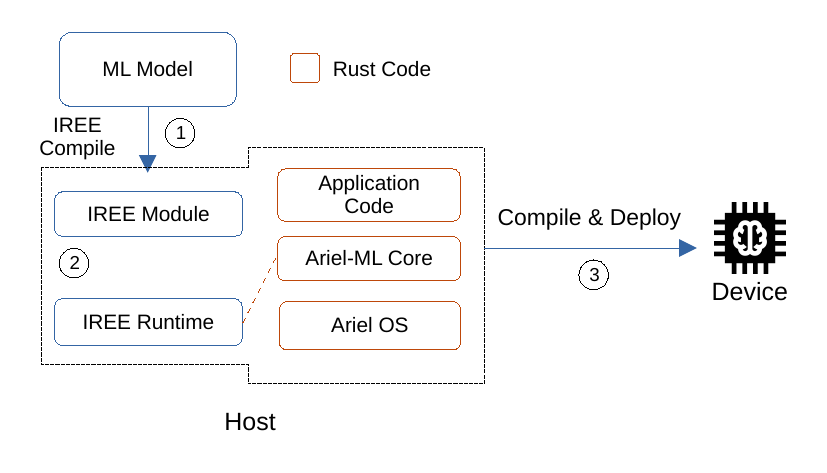}
    \caption{Build pipeline of Ariel-ML. The circled numbers denote the order of execution.}
    \label{fig:build-pipeline}
\end{figure}



Ariel-ML consists of a \textbf{build system} on host and a \textbf{model runtime}  on device. As depicted in \cref{fig:build-pipeline}, the build system executes a build pipeline, where it integrates the IREE compiler to transpile models from mainstream \ac{ML} frameworks (Tensorflow, Torch, ONNX, etc.). The pipeline's workflow contains the following steps:
\begin{enumerate}
    \item Model Compilation. The build system gathers information about the target MCU, including the target triple, MCU features, and alignment requirements, then conveys these details along with the neural network model to the IREE compiler. This process produces an IREE module that encapsulates the model weights, operator implementations compiled into target-specific machine code, and VM bytecode describing the dataflow, invocation sequence, and parallel (tile) configurations for each operator.
    \item Metadata construction. The build system collects and encodes metadata such as the number of available cores and the entry point of the compiled IREE module. This metadata is required by the Ariel-ML runtime during model inference.
    \item Co-compile and Deployment. The IREE module and associated metadata are co-compiled with the application code, IREE runtime, and Ariel OS to produce an executable firmware image, which is then flashed onto the target device.
\end{enumerate}

\begin{figure}[h]
    \centering
    \includegraphics[width=\linewidth]{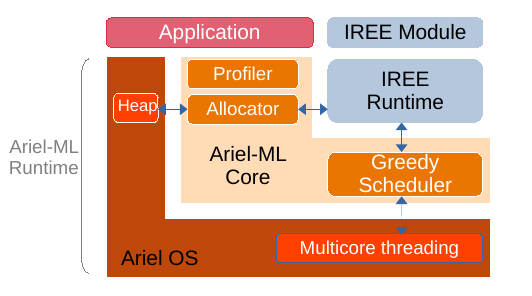}
    \caption{Architecture of Ariel-ML on device.}
    \label{fig:architecture}
\end{figure}

As for the model runtime running on the device side, it is composed of the following key components, as shown in \cref{fig:architecture}.

\begin{itemize}
    \item Ariel-ML core. The Ariel-ML core comprises Rust bindings to the IREE runtime, a profiler for measuring the computational latency and resource utilization of individual operators as well as the overall model, and a greedy scheduling module that dispatches contention-free computational tasks across available cores. It functions as both the control panel for application-level code and the execution backend for model inference. Further implementation details are provided in ~\cref{sec:impl}.
    \item IREE runtime. The IREE runtime is a C-based library that provides essential functionality for invoking models compiled by the IREE compiler. Ariel-ML leverages this runtime to coordinate operator execution and to enable parallel computation within each operator.
    \item OS environment and hardware support. Ariel OS is a Rust-based operating system designed for low-power \ac{IoT} devices. It offers a lightweight runtime environment that facilitates efficient model inference on microcontrollers. This software foundation provides both extensibility and memory safety, ensuring reliable operation across heterogeneous low-power hardware platforms.
\end{itemize}


%% file: sections/05-implem.tex
\section{Implementation Details}
\label{sec:impl}

We implemented Ariel-ML’s multicore scheduler on top of IREE v3.8.0, whose runtime serves as the inference engine responsible for orchestrating the execution of model operators. During model compilation, the computation associated with each operator is partitioned into several contention-free tiles, referred to as \textit{work items}.

At the initialization stage, the Ariel-ML core configures the runtime environment by conveying the model and input data entry points to the IREE runtime, setting up a heap allocator as the memory driver for managing \ac{RAM} resources, and instantiating the greedy multicore scheduler. After this stage, the model is ready for execution.

\begin{figure}[h]
    \centering
    \includegraphics[width=\linewidth]{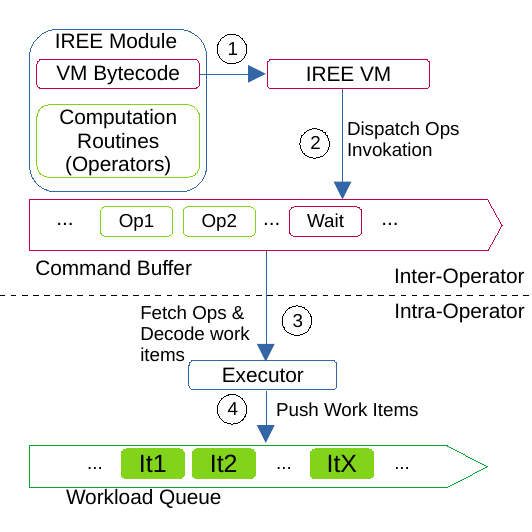}
    \caption{Execution model during model inference. The circled numbers denote the order of execution. \textit{Wait}: synchronization primitive that instructs the executor to suspend until all previously dispatched tasks have finished execution. It: Work Item.}
    \label{fig:exe_model}
\end{figure}

\Cref{fig:exe_model} illustrates the execution model of Ariel-ML. The IREE VM first interprets the bytecode contained in the IREE module, which defines the invocation order, synchronization behavior, and execution semantics of all model operators. These invocations and synchronizations are subsequently transformed into dispatch commands and appended to a command buffer. A command executor then retrieves the operators to be invoked, decodes the associated work items, and places them into a workload queue to be processed by the multicore scheduler.

\begin{figure}[h]
    \centering
    \includegraphics[width=\linewidth]{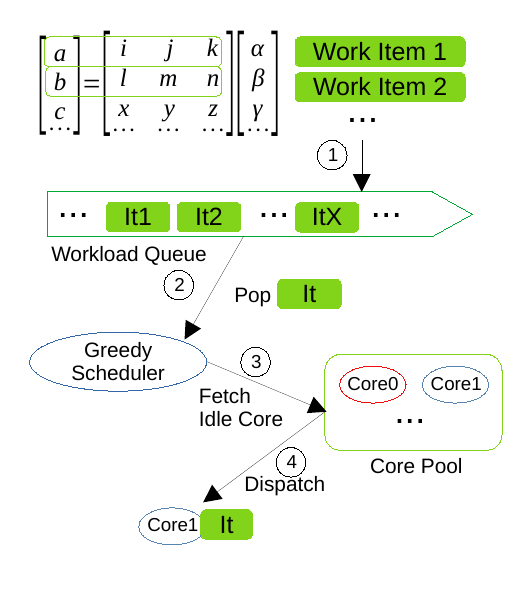}
    \caption{Greedy multicore scheduler in Ariel-ML. Each row–vector multiplication in the matrix-vector multiply operation can be decomposed into conflict-free work items and scheduled across different cores. The circled numbers denote the order of execution. It: Work Item.}
    \label{fig:scheduler}
\end{figure}

\Cref{fig:scheduler} demonstrates the workflow of the greedy multicore scheduler. As work items enter the workload queue, the scheduler continuously pops one item at a time and dispatches it to an available core, repeating this process until the queue becomes empty.

Ariel-ML also provides an optional built-in profiler for benchmarking model execution. When enabled, the profiler records the end-to-end model execution time, per-operator latency, and the corresponding heap and stack usage.

{\noindent\bf Workload Partition and Multicore Scheduling --} 
Ariel-ML relies on IREE to expose schedulable units of work from the input model. During compilation, IREE outlines each operator, such as convolution and dense layers, into an executable entry point, together with parallelizable loops that are further tiled into contention-free regions, referred to as work items. These work items can be scheduled in a lock-free manner across different threads and cores. IREE limits the number of work items of each operator to no more than twice the number of threads to avoid unnecessary context-switching overhead. Furthermore, we set the minimum size of each work item to 80 to further reduce inter-processor communication overhead, as reported in~\cite{ariel-os}.

\section{Correctness Validation of Model Compilation}

We verified the numeric correctness of the compiled models. For each evaluated model, we compared the outputs produced by Ariel-ML against the corresponding reference produced by mainstream \ac{ML} frameworks such as Pytorch and Tensorflow. This evaluation was performed at both the individual operator level and the end-to-end model level. The results show that all deviations are below $10^{-8}$, thereby confirming numerical correctness.

%% file: sections/06-benchmarks.tex
\section{Experimental Results}

In this section, we report on benchmarks we carried out on diverse \ac{IoT}  boards, which are popular representatives of the relevant 32-bit microcontroller families:
\begin{itemize}
    \item {\bf Nordic nRF52840dk}: single-core Arm Cortex-M4 microcontroller at 64 MHz, 1024 kB Flash memory, 256 kB RAM;
    \item {\bf Espressif ESP32C3-devkit}: single-core RISC-V microcontroller at 160 MHz, 384 kB Flash memory, 400 kB RAM;
    \item {\bf RaspberryPi Pico 1}: dual-core RP2040 (Cortex-M0+) microcontroller at 133 Mhz, 2 MB Flash memory, 264 kB RAM.
    \item {\bf RaspberryPi Pico 2}: dual-core RP2350 (Cortex-M33) microcontroller at 150 Mhz, 4 MB Flash memory,  520 kB RAM.
\end{itemize}

{\bf Methodology --} We compare the performance of {\bf Ariel-ML} on heterogeneous hardware against the closest related work to our knowledge. Namely: {\bf RIOT-ML}, which is based on a \ac{TinyML} pipeline combining microTVM and RIOT, implemented in C. We also compare with a hybrid variant we designed and implemented based on a \ac{TinyML} pipeline integrating IREE in RIOT, which we call thereafter {\bf RIOT+IREE}. The rationale is that we are then in a better position to gauge the impact of microTVM {\it versus} IREE for the \ac{TinyML} pipeline, C {\it versus} Rust for embedded code, and single- {\it versus} multi-core for the OS scheduler that is leveraged.

{\bf Experimental Configurations --} For each \ac{MCU}, the core clock frequency was configured to the highest supported value. All programs were compiled with optimization level \emph{O2}. Computational latency was measured by inserting a timer at the beginning of model inference. Each model was benchmarked over ten runs, and the average latency is reported.

{\bf Model Selection --} We selected pre-trained and quantized LeNet-5 \cite{lecun2002gradient} and MCUNet \cite{mcunet-v2} models, designed for typical handwritten digit recognition and visual wake word tasks, respectively. The models' weights and activations are quantized to 8-bit integers, while the input and output tensors remain in IEEE 754 single-precision floating-point format. While for paper limited space reasons we hereafter only present measurements for LeNet-5 and MCUNet, note that Ariel-ML supports a wide variety of models. Please check the supported model zoo at \url{https://github.com/TinyPART/RIOT-ML/tree/main/model_zoo}.

{\noindent\bf Impact of IREE Integration --}
As shown in \cref{tbl:latency}, integrating IREE significantly accelerates model inference. Compared to RIOT-ML, execution times are consistently shorter, regardless of whether the underlying operating environment is C-based RIOT or Rust-based Ariel OS. This improvement comes from two key factors. First, IREE employs more advanced model optimization techniques and a more modern lowering pipeline compared to microTVM. Second, IREE’s embedded LLVM backend enables additional target-specific optimizations during operator code generation, provided that the build system supplies sufficient information about the target \ac{MCU}. These characteristics make IREE particularly suitable for time-sensitive applications in \ac{IoT}  systems.

\begin{figure}
    \centering
    \includegraphics[width=\linewidth, trim={0.8cm 0 0.8cm 0}, clip]{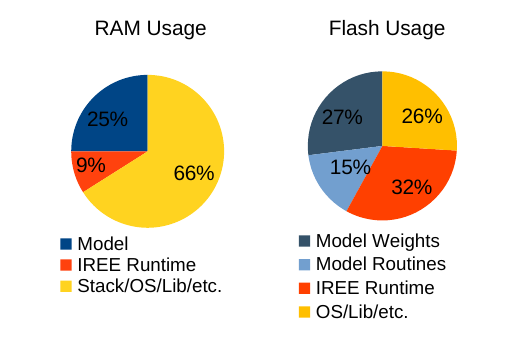}
    \caption{Relative RAM and Flash usage of Ariel-ML with subsystem breakdown on RP2040 (RaspberryPi Pico 1). The model refers to LeNet-5.}
    \label{fig:ram_flash_pie}
\end{figure}

However, this acceleration comes at the cost of increased \ac{RAM} and Flash consumption, as shown in \cref{tbl:ram} and \cref{tbl:flash}. The firmware size increases by approximately $1.1-2.5 \times$ after integrating IREE, while \ac{RAM} usage increases by at least $1.2 \times$. To better understand this overhead, we computed the relative contribution of each component, illustrated in \cref{fig:ram_flash_pie}. For \ac{RAM}, although the IREE runtime itself accounts for only 9\% of the total usage, it requires a 16 kB stack to support the deep function call chain introduced by the IREE VM. In contrast, RIOT-ML requires only a 2 kB stack for model inference. Regarding Flash footprint, the IREE runtime constitutes the largest portion (32\%). 
Moreover, additional Flash footprint is required by FFI (Foreign Function Interface) for Rust interfaces to the IREE runtime, which needs linking against C libraries.

Seen from another angle, if we refer to prior work for LeNet-5 with RIOT-ML (measurements in Table 6 in~\cite{huang2025riot-ml}), we observe that the ML subsystem is approximately 80\%
 of the flash (excluding network stack, SUIT/crypto). Ariel-ML thus remains in the same ballpark, with approximately 75\% of the flash for the ML subsystem for LeNet-5 and equivalent functionality.
Furthermore, despite these overheads, the advanced model optimizations provided by IREE reduce the absolute RAM and Flash usage of the model artifact itself. Overall, there remains substantial optimization room of the IREE runtime to further reduce both RAM and Flash footprints.

{\noindent\bf Impact of Multicore Scheduling --}
Ariel-ML’s multicore scheduling yields a substantial improvement in inference performance. On the dual-core RP2040 (Raspberry-Pi Pico 1) and RP2350 (Raspberry-Pi Pico 2), we observe up to a $1.6 \times$ speedup shown in \cref{tbl:latency}, approaching the hardware’s known limit imposed by bus contention~\cite{ariel-os}. This demonstrates that the greedy, multicore scheduler effectively exploits the available parallelism within the model.

Compared with RIOT+IREE on the same \ac{MCU}, Ariel-ML achieves this speedup with even lower \ac{RAM} usage, while incurring only an 8\% increase in Flash footprint. Compared to RIOT-ML on the same hardware, inference speed tops 200\% with Ariel-ML. 
These results highlight the practicality of multicore-aware scheduling to improve performance on resource-constrained \ac{IoT} devices.

\begin{table}[h]
\caption{Inference execution time (in ms) for quantized LeNet5 and MCUNet. (NA: not applicable; NS: Not supported by OS)}
\centering
\begin{tabular}{llll}
\toprule
MCU        & RIOT-ML & RIOT+IREE & Ariel-ML \\\midrule
\textit{(LeNet5)} \\
nRF52840   & 66.088  & 64.573    & 63.721   \\
ESP32-C3   & 54.953  & 42.138    & 44.17    \\
RP2040     & 70.117  & 50.557    & 46.757   \\
RP2040+multicore & (NA)       & (NA)           & 31.543 ($1.5\times$)   \\
RP2350  & (NS) & (NS) & 27.966 \\
RP2350+multicore & (NS) & (NS) & 19.630 ($1.4\times$) \\\midrule
\textit{(MCUNet)} \\
nRF52840   &  1682.124  &   990.638  &  981.870  \\
RP2040     &  1751.570 &  1055.623   &  1057.223  \\
RP2040+multicore & (NA)       & (NA)           &  661.530 ($1.6\times$) \\
RP2350  & (NS) & (NS) & 396.478 \\
RP2350+multicore & (NS) & (NS) & 280.109 ($1.4\times$)
\\\bottomrule
\end{tabular}
{\scriptsize MCUNet was not benchmarked on ESP32-C3 due to \ac{OOM}.}
\label{tbl:latency}
\end{table}

\begin{table}[h]
\caption{Overall RAM usage (in kB) with quantized LeNet5 and MCUNet. (NS: Not supported by OS)}
\centering
\begin{tabular}{lrrr}
\toprule
MCU      & RIOT-ML & RIOT+IREE & Ariel-ML \\\midrule
\textit{(LeNet5)} \\
nRF52840 & 11.348  & 28.308    & 42.728   \\
ESP32-C3 & 258.874 & 390.154   & 313.192  \\
RP2040   & 28.704  & 44.628    & 42.844   \\
RP2350  & (NS) & (NS) & 47.948\\
\midrule
\textit{(MCUNet)} \\
nRF52840   & 148.284  &  159.884   & 175.680   \\
RP2040     & 164.604 & 176.204 & 174.428  \\
RP2350  & (NS) & (NS) & 166.948\\
\bottomrule
\end{tabular}
\label{tbl:ram}
\end{table}

\begin{table}[h]
\caption{Flash footprint (in kB) with quantized LeNet5 and MCUNet. (NS: Not supported by OS)}
\centering
\begin{tabular}{lrll}
\toprule
MCU      & RIOT-ML & RIOT+IREE & Ariel-ML \\\midrule
\textit{(LeNet5)} \\
nRF52840 & 61.332  & 145.312   & 153.548  \\
ESP32-C3 & 222.272 & 319.042   & 245.320   \\
RP2040   & 65.172  & 160.164   & 172.204  \\
RP2350  & (NS) & (NS) & 172.292\\
\midrule
\textit{(MCUNet)} \\
nRF52840   &  483.580 &  702.860   & 676.080   \\
RP2040    & 484.420  &  725.516   & 782.396   \\
RP2350  & (NS) & (NS) & 693.356 \\
\bottomrule
\end{tabular}
\label{tbl:flash}
\end{table}

%% file: sections/07-discussion.tex

\section{Discussion \& Perspectives}

Our results demonstrate the feasibility of an embedded Rust \ac{TinyML} pipeline supporting heterogeneous multi-core \ac{MCU}-based hardware, which brings benefits regarding model inference latency. At the same time, our study exposes challenges. The use of Rust has an impact on binary size, and interaction between IREE’s execution model and \ac{MCU} architectures introduces new considerations for scheduling, memory layout, and firmware organization. Further refinements are needed (and possible) to fully unlock the potential of deployment of \ac{ML} models in \ac{IoT} devices.

{\noindent\bf Rust Memory Footprint --}
\ac{RAM} and Flash overhead compared to C/C++ equivalents remain substantial, which was also observed in prior works such as~\cite{ayers2022tighten}. On the one hand, we can expect that this overhead will diminish over time as the embedded Rust toolchain is optimized~\cite{sharma2024rust}. On the other hand, as long as the threshold of memory budgets on microcontrollers are not crossed, this overhead has less importance in practice, especially when balanced with the safety advantages that Rust brings over C/C++ in terms of tooling and memory safety~\cite{android-rust}.

{\noindent\bf Size of the IREE Runtime --}
The IREE runtime incurs a large memory footprint due to its complicated, GPU-oriented VM and reliance on the C standard libraries which are necessary on top of the Rust firmware. Future work should explore more feature-tailored build profiles and Rust-native re-implementations of core VM components to substantially shrink the runtime. Such efforts would facilitate IREE's alignment with typical microcontroller memory budgets.

{\noindent\bf Model Structure Support --} 
Ariel-ML inherits its model structure generality from IREE. IREE is not designed around a small collection of neural network templates. Instead, it compiles \ac{ML} models through MLIR-based intermediate representations, including StableHLO, TOSA, and Linalg \cite{lattner2021mlir}. Models from common frameworks such as PyTorch, TensorFlow and ONNX can be imported into this compilation workflow and lowered to executable program routines. Thus, Ariel-ML does not impose structural restrictions such as fixed CNN-wise topologies, which provides support for arbitrary IREE-compatible model structures.

{\noindent\bf On-Device Training Support --}
While Ariel-ML currently targets efficient model inference, emerging \ac{IoT} applications require model fine-tuning or continuous learning on-device. 
One perspective is to enhance Ariel-ML with a lightweight backpropagation mechanism, adapted with IREE compilation and scheduling workflow. This would enable more privacy-preserving, continuously improving \ac{AIoT} applications.

{\noindent\bf Model OTA Update Support --}
Long-term deployment of \ac{AIoT} devices requires reliable, efficient and secure \ac{OTA} model updates. The modular architecture of Ariel-ML and the self-contained nature of IREE modules naturally support differential updates for model weights, operators, or metadata. Developing a secure \ac{OTA} pipeline that leverages this modularity will be essential to maintain model quality, deploy new features, and fix vulnerabilities on fleets of devices in the field.

%% file: sections/08-related-work.tex
\section{Related Work}
{\bf \ac{TinyML} Benchmarking --} 
MLino Bench~\cite{baciu2024mlino} 
provides an open-source benchmarking tool for \ac{TinyML} models on edge devices with limited resources and capabilities.  
MLPerf Tiny~\cite{banbury2021mlperf} defines a set of standard suite of tests for \ac{TinyML} devices. Other works such as \cite{osman2022tinyml-bench} 
or \cite{sudharsan2021tinyml-bench} 
benchmark the performance of TFLM and/or X-CUBE-AI frameworks on some Cortex-M hardware.
Debug-HD~\cite{ghanathe2024debug} 
introduced an on-device debugging approach optimized for KB-sized \ac{TinyML} devices. 
However, no prior work in this area focused on parallelization in multi-core microcontrollers.


{\noindent \bf Microcontroller Hardware-specific Optimizations --} 
Work such as MCUNetV2~\cite{mcunet-v2} or msf-CNN~\cite{msf-CNN} splits the convolution layers of CNNs into partial convolutions to decrease memory requirements on 32bit Cortex-M microcontrollers, at the cost of some re-computation.
Yet other works such as Lupe~\cite{xiang2025lupe} designed an embedded runtime enabling DNN inference on intermittently powered 16-bit microcontrollers (MSP430).  
So far, however, no work has focused on automating computation parallelization for inference on heterogeneous multicore microcontrollers.

{\noindent \bf \ac{TinyML} Model Transpilers \& Compilers --}
Compilers such as TVM~\cite{chen2018tvm}, IREE \cite{The_IREE_Authors_IREE_2019}, FlexTensor \cite{zheng2020flextensor}, and Buddy \cite{zhang2023compiler} provide automated transpilation and compilation pipelines for models developed in major \ac{ML} frameworks, including TensorFlow and PyTorch. As a lightweight extension of TVM, microTVM offers low-level optimizations and runtime support tailored to a variety of processing units, including many microcontroller architectures. 
Tensorflow Lite Micro (TFLM) provides a similar extension for TensorFlow.
More recently, Rust-based transpilers emerged, such as Burn~\cite{burn-rs}, Candle~\cite{candle-rs}, or microflow-rs~\cite{microflow-rs}. However, none of the above supports native use of parallelization on multi-core microcontrollers. 

{\noindent \bf Tiny Machine Learning Operations (TinyMLOps) --}
TinyMLOps extends MLOps to support continuous integration and continuous deployment (CI/CD) workflows of \ac{ML} models on resource-constrained devices. Most MLOps frameworks \cite{kreuzberger_machine_2023, le_tinymlops_2023, diaz-de-arcaya_joint_2023} provide MLOps components for conventional servers or large-scale clusters, but typically do not support \ac{TinyML} devices.
Exceptions are U-TOE~\cite{huang2023u} and RIOT-ML~\cite{huang2025riot-ml}, which do target microcontrollers. However, they do not support multicore -- and they are written primarily in C. 

{\noindent \bf Embedded \ac{IoT} Software Platforms --}
Prior work such as \cite{hahm2015operating} surveys operating systems for microcontrollers. Prominent examples include RIOT~\cite{baccelli2018riot} and Zephyr~\cite{zephyr} which are primarily written in C/C++. Recently, other OS written in Rust have emerged such as Embassy~\cite{embassy}, Ariel OS~\cite{ariel-os} or Tock~\cite{levy2017tock}. However, so far, none of these provide significant support for general \ac{ML} frameworks. The most advanced supports so far are typically hardware- or vendor-specific e.g. with libraries provided by STM32CubeMx or ARM CMSIS-NN. 


%% file: sections/09-conclusion.tex
\section{Conclusions}
The need for a Rust embedded software platform integrating a \ac{TinyML} pipeline has emerged from the simultaneous push for safer systems software building blocks on the one hand, and on the other hand for distributed \ac{AI} more efficiently spread across the Cloud-Edge-Thing continuum. As new microcontroller hardware architectures are increasingly multicore, the ability to fully leverage such capability also becomes necessary. In this paper, we thus introduced Ariel-ML, the first Rust embedded platform automating computation parallelization on heterogeneous multi-core microcontrollers for inference execution using arbitrary \ac{TinyML} models. With the pipeline we designed and the open source toolkit we published, users can transpile, deploy, and execute the model zoo output by various traditional machine learning frameworks (such as PyTorch, TensorFlow) on a wide variety of common \ac{IoT} boards and development kits based on the main microcontroller families (ARM Cortex-M, ESP32, RISC-V). The reproducible benchmarks we provided show how our Rust platform is comparable to prior work in C/C++ on single-core microcontrollers, and how Ariel-ML outperforms prior work on multicore microcontrollers. In a nutshell: Ariel-ML facilitates the tasks of \ac{TinyML} practitioners aiming to gauge feasibility on diverse microcontroller hardware, evaluate the performance of diverse neural network models and, eventually, deploy and run such neural network models in production, integrated on-board in a full-featured embedded Rust operating system. 

\section*{Source Code Availability}
The source code for Ariel-ML is published and maintained at \url{https://github.com/ariel-os/ariel-ml}.

\section*{Acknowledgements}
The work described in this paper was in part financed by France 2030 through the PEPR Future Networks project FITNESS,
and the PTCC project PQ-OTA.